\documentclass[conference]{IEEEtran}
\IEEEoverridecommandlockouts
\usepackage{cite}
\usepackage{booktabs}
\usepackage{amsmath,amssymb,amsfonts}
\usepackage{algorithmic}
\usepackage{graphicx}
\usepackage{textcomp}
\usepackage[T1]{fontenc}
\usepackage{xcolor}
\usepackage{adjustbox}
\usepackage{balance}
\usepackage{enumitem}
\usepackage{subcaption}
\usepackage[utf8]{inputenc}
\usepackage[font=footnotesize]{caption}

\usepackage[a4paper, total={184mm,239mm}]{geometry}
\def\BibTeX{{\rm B\kern-.05em{\sc i\kern-.025em b}\kern-.08em
    T\kern-.1667em\lower.7ex\hbox{E}\kern-.125emX}}

\begin{document}

\title{Autonomous Systems Dependability in the era of AI: Design Challenges in Safety, Security, Reliability and Certification}

\author{\IEEEauthorblockN{Behnaz Ranjbar\IEEEauthorrefmark{1},
Kirankumar Raveendiran\IEEEauthorrefmark{2}, 
Sudeep Pasricha\IEEEauthorrefmark{2}, 
Samarjit Chakraborty\IEEEauthorrefmark{3},\\
Cecilia Carbonelli\IEEEauthorrefmark{4}, 
and~Akash Kumar\IEEEauthorrefmark{1}
}
\IEEEauthorblockA{\IEEEauthorrefmark{1}Chair of Embedded Systems, Ruhr Universität Bochum, Germany}
\IEEEauthorblockA{\IEEEauthorrefmark{2}Colorado State University, US}
\IEEEauthorblockA{\IEEEauthorrefmark{3}The University of North Carolina at Chapel Hill, US, and The TUM Institute for Advanced Study, Germany}
\IEEEauthorblockA{\IEEEauthorrefmark{4}Infineon Technologies, Germany}

\thanks{This is an extended version of the focus session paper published in Design, Automation \& Test in Europe Conference \& Exhibition (DATE) 2026.}
}


\maketitle

\begin{abstract}
The design of embedded safety-critical systems such as those used in next-generation automotive and autonomous platforms, is increasingly challenged by escalating system complexity, hardware–software heterogeneity, and the integration of intelligent, data-driven components. Ensuring dependability in such systems requires a holistic approach that spans multiple abstraction layers and encompasses both design- and run-time assurance. Traditional methods for reliability, safety, and security management often fall short in addressing the dynamic and uncertain behaviors introduced by Artificial Intelligence~(AI) and Machine Learning~(ML) components, especially under stringent real-time, power, and safety constraints. While AI and ML offer powerful predictive, adaptive, and self-optimizing capabilities that can enhance system dependability, their inherent non-determinism, data-dependence, and lack of formal guarantees introduce new challenges for verification, validation, and certification. This paper explores emerging methodologies, architectures, and frameworks for designing dependable autonomous and embedded systems in the era of AI. It highlight advances in reliability modeling, secure system design, and certification approaches that account for imperfect, learning-enabled components, aiming to bridge the gap between AI innovation and certifiable system-level dependability. 
\end{abstract}


\section{Introduction}
The deployment of embedded safety-critical systems for next-generation automotive and autonomous platforms has become increasingly significant in recent years. This growth arises rapidly escalating system complexity, the increase in carrying heterogeneous hardware–software architectures, and the integration of intelligent, data-driven components~\cite{machines13080646,Liu2021}. While these components enhance functionality and autonomy, they pose significant challenges for predictability, power efficiency, security, reliability, and safety assurance. To this end, existing design, verification, and certification methodologies are often insufficient to address the needs of modern safety-critical automotive systems. 

\begin{figure}
    \centering
    \includegraphics[width=0.99\linewidth]{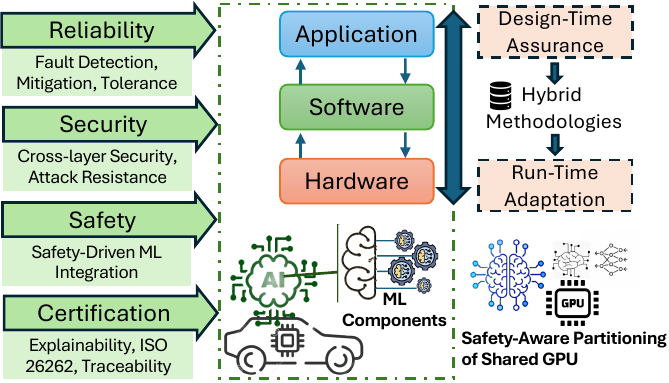}
    \caption{An Overview of Autonomous System Dependability Management}
    \label{fig:overview}
\end{figure}

Dependability in these systems can be achieved through methodologies that consider multiple abstraction layers and address both design- and run-time phases. 
While design-time approaches aim to guarantee correct execution with reliability and safety constraints before deployment, run-time approaches are also essential to monitor system behavior, handle uncertainty, and ensure dependability under unexpected behavioral changes~\cite{Ranjbar2023,sahoo2021reliability}. Fig.~\ref{fig:overview} demonstrates the overview of dependability management across abstraction layers for Autonomous Vehicle~(AV) at both design- and run-time. 

Traditional approaches to managing reliability, safety, and security are often inefficient when faced with the deployment of Artificial Intelligence (AI) and Machine Learning (ML), due to the uncertain behaviors they introduce. 
AI and ML exhibit data-dependent and adaptive behaviors, which are difficult to model and verify in safety-critical systems with strict requirements~\cite{Hegde2025}. The challenges are more significant when considering the stringent real-time, power, and safety constraints imposed on these safety-critical embedded systems.
While the deployment of AI and ML in autonomous system design provides predictive, adaptive, and self-optimizing capabilities that can enhance system dependability, their inherently non-deterministic, data-driven, and lack of formal guarantees create new verification, validation, and certification challenges. As a result, it motivates the need for novel assurance methodologies.

In this paper, methodologies, architectures, and frameworks are explored for the design of dependable autonomous systems in the era of AI. It highlights reliability models, secure system design, and certification approaches for imperfect and ML components. By addressing the limitations of existing assurance techniques, the paper aims to bridge the gap between AI and the need for stringent and certifiable system-level dependability.

The paper is organized as follows. We first examine how ML techniques can help to achieve cross-layer reliability improvements in autonomous systems, while highlighting the key challenges associated with ML deployment (Section~\ref{sec:CLRAV}). Section~\ref{sec:CLSAV} then discusses cross-layer security in automotive systems, with particular emphasis on the use of ML-based mechanisms. Section~\ref{sec:SAFEAV} addresses the problem of designing safe autonomous systems in the presence of imperfect ML components, discussing strategies to mitigate their limitations and uncertainty. Finally, open challenges related to reliability and security in AI-based sensing systems for safety-critical applications are discussed in Section~\ref{sec:SecureAI}.

\section{Cross-Layer Reliability for Safety-Critical Automotive Systems}
\label{sec:CLRAV}

A wide range of embedded systems found in many industrial application domains, such as automotive, are safety-critical. In these systems, various applications with different assurance levels are executed onto a platform while operating under stringent constraints on cost, space, timing, and power consumption, and guaranteeing a safe operation, simultaneously. In these applications, tasks are classified into multiple safety levels to maintain the predictability of the applications under different unexpected behaviors~\cite{Wu2020,sahoo2021reliability}. The classification is done based on the functional importance of tasks to overall system operation. ISO 26262 is an industrial safety standard for automotive applications and defines functional safety levels, known as Safety Integrity Levels (SIL)~\cite{Ernst2016} ranging from ASIL-A (lowest) to ASIL-D (highest)), shown in Table~\ref{tab:SafetyStandarad}. Failures impact systems differently depending on the assigned levels, and the corresponding Probability of Failure per Hour (PFH) is specified accordingly.

In the automotive industry, AVs represent one of the most advanced forms of embedded safety-critical systems, operating under strict real-time constraints, serve increasingly as safety-critical components and managing vast heterogeneous data. Light Detection and Ranging (LiDAR)-based application is a key component of autonomous driving systems and, consists of several main modules, which contains highly automated driving functions: (1) Raw data processing, which involves analyzing and processing sensor data for object detection; (2) Perception, responsible for identifying, classifying, detecting, and tracking the object, such as object detection, lane detection, road sign recognition using sensors; (3) Localization, which determines the position and orientation of the vehicle; (4) Prediction, where the system predicts the future behavior of surrounding agents, such as pedestrians, vehicles, and cyclists; (5) Planning, which generates a safe path for the vehicle based on perceived information and predictions, to avoid objects and obstacles; (6) Vehicle control, which converts planned information into precise vehicle control signals such as steering, braking, and acceleration. All these tasks must be executed continuously with strict timing constraints corresponding to their safety level.

\begin{table}[t]
    \centering
    \caption{ISO 26262 safety requirement~\cite{ISO26262}.}
    \vspace*{-.1cm}
    \label{tab:SafetyStandarad}
    \adjustbox{max width=\linewidth}{
    \begin{tabular}{ccccc}
    \hline
         $\zeta$ & ASIL-D & ASIL-C & ASIL-B & ASIL-A \\
         \hline
         \hline
         $PFH(\zeta)$ & $10^{-8}$ & $10^{-7}$ & $10^{-6}$ & $10^{-5}$ \\
         \hline
         Failure Condition & Catastrophic & Hazardous & Major & Minor\\
         \hline
    \end{tabular}
    }
    \vspace*{-.2cm}
\end{table}

\subsection{Importance of Reliability Assurance and Challenges}

Unlike traditional automotive systems, autonomous systems operate with less human supervision, and are responsible for real-time environment interpretation, safe planning, and execution of vehicle control functions dynamically. Any faults, delayed response, or incorrect decision in doing these modules may lead to catastrophic consequences. Moreover, due to existing numerous sensors, interfaces and software components, it increases the possibility of unexpected interactions, hidden dependencies, and rare operational failures that traditional testing methods cannot fully uncover. Achieving high reliability in autonomous systems is complicated by the proliferation of fault sources across software, and application layers.

In addition, while the complexity of these applications has been increased, the technology scaling in these modern embedded platforms has dramatically exacerbated the rate of manufacturing defects and physical fault rates. Therefore, the safety and reliability issues have increased tremendously across all electronic systems. Increased unreliability, both in terms of computation errors and the reduced life-time of systems, has led to the increasingly complex problem of ensuring the reliable execution of applications on increasingly unreliable hardware~\cite{sun2019dynamic,sahoo2021reliability}. Since failure in such safety-critical systems may cause catastrophic consequences, improving reliability under all circumstances of stress and environmental changes plays a vital role in these systems~\cite{sahoo2021reliability,singh2013mapping}.

\begin{figure}
    \centering
    \includegraphics[width=0.99\linewidth]{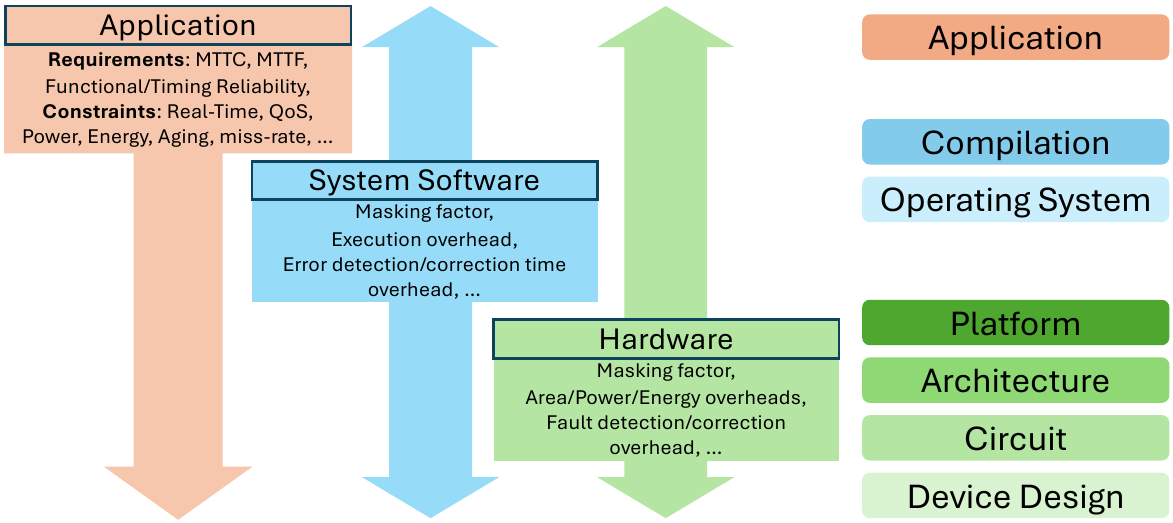}
    \vspace*{-.3cm}
    \caption{Cross-Layer Reliability-Aware System Design}
    \vspace*{-.2cm}
    \label{fig:CLFaultT}
\end{figure}

Since faults may propagate through hardware and software layers, cross layer reliability analysis is essential. In order to design a reliable system, fault mitigation and reliability methods need to be applied in multiple system abstraction layers, like hardware, application, and system software~\cite{sahoo2021reliability,Ranjbar2023}. The single-layer reliability-aware design methodologies, like those in~\cite{Navardi20222,ranjbar2020fantom,Ranjbar2022,Ranjbar2023}, adopt an other-layer-agnostic approach~\cite{sahoo2019hybrid}. However, this isolated layer-wise fault-mitigation has a high-cost (in terms of power, area, and timing), making it infeasible for most embedded systems~\cite{sahoo2016cross,sahoo2021reliability}. Therefore, cross-layer solutions, like what presented in~\cite{sahoo2020cl,sahoo2019hybrid}, are applied to provide application-specific and low-cost fault-tolerance by distributing the fault mitigation activity across the layers. It can involve distributing fault-mitigation activities to several layers, resulting in a reduced fault-mitigation effort at the hardware layer leading to more cost-effective designs. This power cost can also be improved through low power techniques, such as those presented in~\cite{ranjbar2023power}. A comprehensive overview of fault-tolerant methods was provided in~\cite{sahoo2016cross,sahoo2023fault} and their suitability for cross-layer design was discussed. Fig.~\ref{fig:CLFaultT} shows a reliability-aware system design across three layers -- Hardware, Software, and Application -- for embedded real-time systems. This information is essential for evaluating the suitability of fault tolerance techniques for an application and effect of their overheads at other layers, to improve reliability. 

Design Space Exploration (DSE) for these solutions is required to not only select the mechanism at each layer but also their appropriate configurations, leading to a huge design space~\cite{Sahoo2021}. Moreover, it is infeasible to analyze all possible scenarios of system design at design-time to prevent unexpected behavior that may occur at run-time. Further, these approaches are investigated at design-time to guarantee the reliability of embedded safety-critical systems in the worst-case scenario. However, the worst-case scenario rarely occurs at run-time. Appropriate run-time approaches are needed to maintain and improve reliability. To this end, modeling and analyzing cross-layer reliability, and designing practical cross-layer DSE approaches are required to manage reliability at both design- and run-time to reduce overheads and complexity. 

To further discuss the challenges, note that autonomous systems require proactive reliability improvement methods capable of predicting faults before occurrence, preventing hazardous situations, and ensuring continuous safe operation. Achieving this objective presents several key challenges such as insufficient testing coverage, high system complexity which makes system behavior difficult to predict and improve safety and reliability, and difficulty in early fault detection in which, without intelligent monitoring mechanisms, many faults are detected after failure has occurred. 

To this end, these challenges collectively motivate the adoption of ML–based approaches to enhance reliability assurance in autonomous embedded systems.

\subsection{ML for Reliability Assurance: Benefits and Open Challenges}

ML techniques offers several unique capabilities that effectively address the limitations of traditional reliability management mechanisms. 
First, from the perspective of cross-layer reliability improvement, several system-level design- and run-time approaches have been exploited. However, existing management approaches like~\cite{sahoo2019hybrid} are not adaptive to varying system conditions, unexpected requirements, and workload variations during run-time or need prior knowledge of the system and applications, leading to large profiling overheads. In fact, to maximize the benefits from such approaches, the system needs to dynamically adapt to changes in the operating environment and reduce the complexity. Hence, ML techniques are suitable for adaptive run-time operation including fault prediction, detection, and tolerance. These techniques can learn from past events, gain knowledge, and improve the model, to determine the ideal system configuration under environmental changes. Besides, due to the large design complexity of implementing cross-layer reliability, ML techniques can help and are employed to determine efficient techniques to detect and predict, and tolerate faults in these safety-critical systems.

However, employing ML can introduce new challenges in ensuring requirements and trustworthiness. They introduce significant overheads; ML-based cross-layer reliability models must trade accuracy against computational and energy costs, and battery life. Therefore, analyzing ML techniques and identifying the suitable ones for AV are still needed to be studied.
In addition, there are significant challenges in using ML for reliability assurance of automotive safety-critical applications: 
\begin{itemize}
    \item Lack of explainability. ML often behave as functionally opaque model, making it difficult sometimes to justify decisions to guarantee real-time and safety constraints.
    \item Data limitation: ML should rely on high-quality, diverse, and representative datasets. Rare failures or unsafe scenarios are underrepresented, which reduces model reliability for critical cases.
    \item ML models may perform poorly under unseen conditions. Dataset biases can cause unsafe misclassifications. 
    \item ML models require real-time decision-making in automotive applications. Large models may exceed timing constraints that may reduce safety levels.
    \item ML errors may predict wrong and directly result in hazardous behavior. Reliability must therefore be considered both in terms of performance and failure impact.
\end{itemize}

\subsection{Paths Forward}

In future work, identified challenges must be addressed through the development of effective solutions. Moreover, future research should investigate interpretable ML models for adaptive reliability management and establish standards for ML-based reliability assurance across hardware, software, and application layers. In addition, ML models should be deployed at run-time to support continuous reliability management and enhance life-time reliability. The development of self-aware and self-healing automotive systems, capable of automatically detecting and tolerating faults and improving reliability by combining different techniques in different hierarchical layers, also represents a promising research direction. 

\section{Cross-Layer Security for Safety-Critical Automotive Systems}
\label{sec:CLSAV}
The increasing connectivity in emerging AVs has led to a surge in potential cyberattacks and expanded the automotive attack surface, introducing vulnerabilities at multiple system layers~\cite{Kukkala2022}. High-profile attacks, such as the remote attack on an unaltered 2014 Jeep Cherokee~\cite{miller2015remote} enabled remote control of critical functions and highlight how security needs to be an important design concern alongside safety and real-time performance. Ensuring robust security in safety-critical AVs therefore requires a comprehensive cross-layer approach: securing low-level in-network communications and ECU firmware, actively monitoring in-vehicle networks for intrusions, securing perception, localization, and control modules, and ensuring attack-free connected vehicle (V2X) scenarios.

\subsection{In-Vehicle Network Security}

At the lowest layer in the AV stack, ensuring message authenticity and confidentiality on the in-vehicle network is paramount. However, adding cryptographic protection to safety-critical messages is non-trivial, as encryption and authentication operations consume CPU time in ECUs and can delay time-sensitive communications. In resource-constrained ECUs, naively applying strong cryptography may overload processors and cause deadline misses. 

A promising solution to this challenge is the cross-layer SEDAN framework~\cite{Kukkala2020}, which was developed to maximize security without violating real-time deadlines for time-triggered automotive networks. At design time, SEDAN allocates tasks to available ECUs in the system and generates the set of signals needed for inter-task communication. These signals are packed into messages using a frame packing approach, and security requirements are derived for each message. The size of the keys used for encryption and decryption of the messages are optimized using a greedy randomized adaptive search procedure~(GRASP) meta-heuristic. At runtime, SEDAN incorporates lightweight cryptographic primitives using efficient symmetric-key ciphers over expensive asymmetric key cryptography and a novel meta-heuristic key management scheme to distribute cryptographic keys across ECUs to perform authenticated encryption and decryption of messages. A runtime scheduler then makes use of the generated keys and the optimal design time schedule to schedule messages via a fast heuristic. SEDAN intelligently tunes security per message based on criticality by considering the automotive safety standard ISO 26262 and real-time performance. Experiments on a real vehicle dataset derived from a Chevrolet Camaro showed that SEDAN meets all safety-critical deadlines while significantly improving security, surpassing state-of-the-art efforts. 

Even with cryptographic protections in place, no system is entirely impenetrable – sophisticated attackers may find ways to infiltrate the vehicle’s network via an infected ECU, diagnostic port, etc. Thus, an Intrusion Detection System~(IDS) that continuously monitors the in-vehicle network is crucial. Automotive IDS solutions generally fall into two categories: signature-based (looking for known attack patterns) and anomaly-based (looking for deviations from learned normal behavior). Signature-based IDS can detect known attacks quickly with low false alarms, but they fail against novel or unforeseen attacks. In contrast, anomaly-based IDS can potentially flag new attack vectors (including insider attacks or zero-days) by learning what “normal” operation looks like and detecting deviations. Given the impracticality of enumerating all possible automotive attacks and the availability of abundant normal driving data (through testing and simulation), anomaly-based IDS has emerged as the more practical approach for in-vehicle networks. 

Several promising AI-driven approaches have been proposed that can enable efficient IDS. INDRA~\cite{Kukkala2020a} is an IDS that learns the vehicle’s normal network behavior in an unsupervised manner using a uses a Gated Recurrent Unit (GRU) based recurrent autoencoder and then detects deviations indicative of attacks. During an offline training phase, the GRU model is trained to reconstruct sequences of normal CAN network messages, effectively capturing temporal patterns and correlations in legitimate data. At runtime, INDRA computes an intrusion score for each new message sequence based on the reconstruction error. If the framework cannot reliably reproduce the observed sequence, it signals an anomaly. Across various attack scenarios (e.g., continuous, plateau, playback, suppress, flooding), INDRA achieved high detection rates (> 80\% for many attacks) and a false alarm rate under 1\%. LATTE~\cite{kukkala2021latte} further improved upon INDRA’s detection capabilities, with a Long Short-Term Memory (LSTM) encoder-decoder, augmented with a self-attention layer that learns to weigh the LSTM’s hidden states to help the model focus on critical time steps or signal IDs that strongly indicate the system’s state, filtering out noise or routine fluctuations. LATTE delivered superior accuracy and F1-score, and fewer false positives than INDRA and other state-of-the-art methods across a range of attack types. Its efficient design, with ~1.4MB model size and ~193µs inference latency, means that LATTE can run in real-time, processing incoming message frames on-the-fly within the short inter-message intervals of automotive networks. TENET~\cite{thiruloga2022tenet} further improves on this framework with a Temporal Convolutional Neural Network (TCN) and an integrated attention mechanism. Compared to INDRA, TENET achieved 3\% higher detection rate including a 17.3\% improvement in ROC-AUC and a 32.7\% reduction in false negative rate. Importantly, TENET’s model has 94.6\% fewer parameters than INDRA and runs with 48.1\% lower inference latency on the same embedded platform. Such frameworks can ensure that an attacker who manages to inject malicious packets will have a high chance of being detected, allowing the vehicle to trigger fail-safes or alerts before safety is compromised.

\subsection{In-Vehicle Perception Security}

Modern AVs rely on an array of sensors such as cameras, LiDARs, radar, ultrasound, etc. to perceive their environment. The data from these sensors feed into object detection, tracking, and planning algorithms that make split-second driving decisions. Anomalies in this perception pipeline due to malicious manipulation have dire safety consequences. For example, an attacker might deliberately induce perception anomalies (e.g., placing doctored road signs), leading the vehicle to make dangerous decisions. Thus, robust anomaly detection at the sensor and object detection layer is important. 

Modern AI-driven anomaly detection approaches in AV perception can be categorized as reconstruction-based, prediction-based, or confidence-based methods. Among these, reconstruction-based strategies have gained traction by leveraging deep generative models to flag out-of-distribution sensor data via reconstruction error. As an example, \cite{Sboui2024} targets anomalies in LiDAR spatio-temporal data and argues that labeled anomaly data is expensive and impractical, motivating unsupervised detection. This work proposes a CNN–BiLSTM variational autoencoder (VAE) architecture: CNN layers learn spatial structure while BiLSTMs capture temporal dependencies in both directions. Evaluation on the Lyft motion prediction dataset with injected obstacle-like anomalies to mimic real conditions revealed a test accuracy of 90.2\% compared to 85.1\% of a state-of-the-art method. Most importantly for embedded deployment, it reports substantially lower memory footprint than the state-of-the-art (98KB vs. 1.14MB) while improving reconstruction quality. However, a known challenge is that powerful autoencoders might accidentally reconstruct anomalies as if they were normal inputs, reducing detection efficacy. To address this, GAAD~\cite{Rezaei2024} integrates a generative adversarial network (GAN) with an autoencoder. GAAD’s generator-autoencoder module is trained to reproduce only the distribution of normal sensor data, while the discriminator learns to distinguish genuine (normal) sensor readings from anomalous ones. This GAN–VAE hybrid enables real-time detection using fixed-width sliding window over multi-sensor streams to create a continuous input during a trip. On the Lyft Level 5 prediction dataset, GAAD achieving AUROC scores above 0.98 across varying anomaly types, while improving reconstruction accuracy over the state-of-the-art. Such approaches to securing the perception layer will be increasingly important with the greater reliance on efficient perception in AVs.

\subsection{Vehicle Localization and Mapping Security}

Accurate localization is foundational for AVs, where the vehicle must know its position and orientation to stay in lanes, avoid obstacles, and navigate to destinations. AVs localize using a combination of GPS/GNSS, inertial sensors, and environmental mapping comparing camera/LiDAR data to high-definition maps. Each of these components has security weaknesses that are being increasingly scrutinized. 

GPS signals, for example, are weak and unencrypted. Adversaries can mount GPS spoofing attacks by broadcasting counterfeit GPS signals that cause a vehicle’s receiver to calculate a wrong position. In one high-profile study, researchers showed that under certain dynamic conditions like when a vehicle is accelerating or cruising at high speed, GPS spoofing becomes easier and can quickly steer an AV off its intended course~\cite{zhang2025ghost}. This kind of attack, essentially a “ghost navigator” feeding the car a false trajectory–highlights that even redundant sensors can be fooled if the attack is cleverly synchronized with vehicle motion. 

Defenses for localization attacks often leverage cross-layer sensor verification. A simple mitigation for GPS spoofing is redundancy: use multiple GNSS constellations (GPS, GLONASS, Galileo) and reject solutions that are inconsistent between them. More sophisticated approaches use the vehicle’s perception sensors to double-check location: for instance, camera-based lane detection or recognizing landmarks (signs, buildings) can provide an independent position estimate to compare against GPS. Another line of defense is anomaly detection on the localization output: if the computed position jumps erratically or contradicts the vehicle’s own motion model (e.g., requiring an acceleration beyond physical limits), the system can raise an alert or fall back to a safe-state like prompting the human driver to take over. A recent work developed an anomaly detector for GPS that uses AI to identify subtle spoofing attempts which mimic normal noise, by analyzing the sequential consistency of location readings~\cite{abrar2024gps}.

Beyond localization, the mapping process itself can be attacked. AVs rely on stored maps of the environment; an attacker who hacks the map data either on the vehicle or at the server providing maps could introduce phantom obstacles or remove real ones. For example, falsifying a map to erase a known one-way street or add a nonexistent road could cause dangerous routing decisions. Protecting map data involves both cryptographic integrity and consistency checks involving comparing sensor observations against the map to detect discrepancies. The use of blockchain or distributed ledgers is actively being explored to ensure trust in crowd-sourced map updates, though such solutions are still experimental

\subsection{Vehicle Control Layer Security}

At the control layer, the vehicle’s computing system turns sensor inputs and high-level plans into physical actions (steering, throttle, braking). These controllers are typically designed with built-in safety monitors (for instance, to prevent engine damage or ensure stability), but historically they assume the inputs from sensors or driver commands are trustworthy. A clever attacker who infiltrates the system could exploit this by injecting false signals or command values that, while plausible, lead to dangerous behavior. One canonical example is false data injection into a Cooperative Adaptive Cruise Control (CACC) system. In CACC, vehicles wirelessly share their speeds and intentions to maintain a tight platoon; if an attacker spoofs a leading vehicle’s data or distorts a follower’s sensor readings, it can cause the control algorithm to misjudge distances and potentially trigger collisions. 

Researchers have begun developing attack-resilient control algorithms to address such threats. A secure observer-based control approach for CACC that can tolerate a certain level of erroneous data was proposed in~\cite{bonab2025secure}. Their approach uses a Luenberger observer augmented with anomaly detection: by comparing the observer’s predicted vehicle state with the actual sensor state, the controller can tell if sensor inputs have been corrupted by an attack. They also apply Lyapunov stability analysis to prove that even under sustained false-data attacks, the platoon will remain stable and safe as long as the attack is bounded in magnitude. This kind of robust control logic is a form of cross-layer defense: it operates at the control software layer but is deeply informed by physical dynamics (e.g., vehicle braking distances, engine response). 

Another facet of control security is ensuring the integrity of the controller code and firmware. Techniques like secure boot and runtime memory safety monitors come into play here. Each ECU that governs braking or steering should verify its software has not been tampered via cryptographic signatures each time it powers on. If an attacker somehow manages to alter the control logic (e.g., via a firmware update exploit), a secure boot will detect the invalid signature and prevent the compromised module from running. Research on hardware-supported security for AV controllers (e.g., using trusted execution environments) is ongoing. The overarching goal is graceful degradation: even if part of the system is compromised, the vehicle should fail safe~(e.g., slow to a stop) rather than behave erratically. Ensuring this requires tight coordination between the control software and low-level hardware, as well as cross-layer awareness.

\subsection{Connected Vehicle (V2X) Security}

Beyond the confines of a single vehicle, AVs will increasingly communicate with external entities: other vehicles~(V2V communication for collision avoidance and platooning), roadside infrastructure (V2I communication with traffic signals or cloud servers), and personal devices. This greatly expands the attack surface – now an attacker does not need physical access or to already be on the in-vehicle network; they can potentially exploit wireless channels (DSRC, C-V2X, Wi-Fi, Bluetooth, cellular) to orchestrate attacks. The automotive industry and standards bodies have anticipated some of these risks: V2X communication standards like IEEE 1609.2 (WAVE) mandate the use of digital signatures on V2V messages and a Public Key Infrastructure for vehicles. However, even with cryptographic authentication, certain attacks remain possible – notably Sybil attacks, where an adversary generates many false vehicle identities each with valid credentials, to, say, flood an area with fictitious traffic congestion reports. 

Researchers have surveyed the landscape of connected vehicle threats and defenses~\cite{dibaei2020attacks}, categorizing them into attacks on privacy (eavesdropping on V2X messages or tracking vehicles), authenticity (broadcasting false information), and availability~(jamming or denial-of-service on wireless channels). As vehicles move toward higher autonomy and cooperation~(e.g., coordinated intersection passing), ensuring the fidelity of V2X messages is as important as securing the in-car systems. Defenses will need to span multiple categories: cryptographic protocols~(to ensure confidentiality and authenticity), misbehavior detection systems (to identify nodes that are sending implausible or inconsistent data), and network security techniques~(firewalls, intrusion detection on the vehicular cloud).


\section{Designing Safe Autonomous Systems with Imperfect ML Components}
\label{sec:SAFEAV}

Modern cyber-physical systems (CPS)~\cite{GoSMLCAV12} increasingly rely on deep neural networks (DNNs) for perception and state estimation, replacing traditional model-based observers in tasks such as object detection, localization, and environment understanding. While DNN-based perception has enabled major advances in autonomy~\cite{BoCJJR23}, it also introduces two fundamental sources of uncertainty that directly impact system safety. First, inference uncertainty arises because practical CPS platforms are resource constrained: power, memory, and compute limitations often prevent deployment of large, highly accurate DNNs on edge devices, necessitating the use of smaller and less accurate models. Second, offloading perception workloads to cloud or remote accelerators enables higher accuracy but introduces additional sensing-to-actuation delays due to communication and computation latency~\cite{FrJXJSJC23,GhHXSAJTBBDC24}. Both inference errors and delays propagate through feedback control loops and can significantly degrade closed-loop performance, even when individual components appear correct in isolation \cite{OeBBC14,GoSC11,SaAWLSKC14}. This creates a key challenge for dependable autonomous systems: how to design CPS that remain safe and certifiable despite imperfect ML components embedded in perception pipelines~\cite{TiMSNZC16}. In the remainder of this section, we illustrate two concrete, system-level design approaches that explicitly address this challenge.

\subsection{Safety-Driven Edge--Cloud Partitioning of Perception DNNs}

The first example addresses the problem of designing safe control systems when perception is performed using a combination of edge and cloud DNNs~\cite{CaXFFC24}. In many autonomous systems, executing perception entirely on the edge leads to low latency but high estimation error~\cite{CaFFCF25}, while executing perception entirely in the cloud yields higher accuracy at the cost of significant delays. Naively choosing either option can result in unsafe transient behavior, even if the underlying controller is stable \cite{hobbs2022,ScGZLC11}. The key insight here is that safety depends on the \emph{combined effect} of estimation uncertainty and delay, rather than on either factor alone.

To address this, we modeled perception and control as an integrated system and introduced a hybrid edge--cloud control strategy. A lightweight DNN deployed on the edge produces fast but noisy state estimates, enabling low-latency control actions, while a larger and more accurate DNN running in the cloud produces delayed but higher-quality estimates. Instead of treating these two estimates independently, the controller is explicitly designed to combine them in a structured way, accounting for their different delays and uncertainty characteristics. Such a control architecture that adapts its actuation strategy as more accurate information becomes available, rather than assuming perfect or instantaneous perception \cite{zhu2025,ChS25}.

Crucially, the design process does not rely on idealized assumptions such as zero delay or perfect estimation. Instead, it quantifies how inference uncertainty and cloud latency affect system trajectories and evaluates safety using system-level metrics such as deviation from a nominal trajectory. This enables principled exploration of the edge--cloud design space, including how much of the DNN should be placed on the edge versus the cloud, and how controller gains should be chosen to compensate for delayed information \cite{ChS25,PZWGXDC25}. Experimental results on autonomous vehicle benchmarks show that appropriately sizing and partitioning edge and cloud DNNs can significantly reduce trajectory deviation and control effort compared to purely edge-only or cloud-only solutions, yielding both improved performance and stronger safety margins.

\subsection{Safety-Driven GPU Partitioning and DNN Sizing}

The second example considers a complementary problem: how to design safe CPS when \emph{multiple} DNN-based perception tasks share limited on-board GPU resources. In many autonomous platforms, several neural estimators run concurrently, each responsible for estimating a different component of the system state (e.g., position, velocity, orientation). A common design practice is to treat all perception tasks uniformly, allocating similar network sizes or compute budgets to each estimator. However, we have demonstrated that this approach can be inefficient and unsafe, because errors in different state variables do not affect closed-loop safety equally \cite{ZLGCD25}.

The proposed solution introduces a safety-driven GPU partitioning methodology that explicitly accounts for the \emph{sensitivity} of system-level safety metrics to estimation errors in different state components. Instead of optimizing individual DNN accuracy in isolation, the approach evaluates how uncertainty in each estimated variable propagates through the control loop and affects reachable sets of the closed-loop system. This enables prioritizing computational resources for perception tasks that have the greatest impact on safety~\cite{HoXFDDC24}, while assigning smaller models to less critical variables. To make this approach practical, we have developed several heuristics~\cite{XuHSAGYSHJHDC24} that efficiently explore the tradeoff between GPU cost~\cite{CaFFSPCF24} and system-level safety without exhaustive search. These include dynamic programming-based exploration, fast greedy allocation strategies, and sensitivity-guided heuristics that require only a small number of reachability computations \cite{XuHGZASYSHJHDC24}. 

To understand these approaches in more detail, consider a closed-loop cyber--physical system with discrete-time dynamics
$
    x_{k+1} = f(x_k, u_k, w_k),
$
where $x_k \in \mathbb{R}^n$ denotes the system state, $u_k$ is the control input, and $w_k$ captures bounded disturbances. The control input is generated using state estimates $\hat{x}_k$ produced by a set of DNN-based perception modules, each estimating a subset of state components. Due to resource constraints, these estimators introduce bounded estimation errors, which we model as:
$
    \hat{x}_k = x_k + e_k,$ with $e_k \in \mathcal{E}(\boldsymbol{\theta}),
$
where $\mathcal{E}(\boldsymbol{\theta})$ is a parameterized uncertainty set whose size depends on the DNN architecture configuration $\boldsymbol{\theta}$, such as network depth, width, or allocated GPU resources.

Safety is specified in terms of a state constraint $x_k \in \mathcal{X}_{\text{safe}}$ that must hold for all time. Given a control policy $u_k = \pi(\hat{x}_k)$, the effect of estimation uncertainty is analyzed through reachability: for a fixed allocation $\boldsymbol{\theta}$, we compute an over-approximation of the reachable set:
\vspace*{-.2cm}
\begin{equation*}
    \mathcal{R}_{k+1}(\boldsymbol{\theta}) = \left\{ f(x, \pi(x+e), w) \;\middle|\; x \in \mathcal{R}_k,\, e \in \mathcal{E}(\boldsymbol{\theta}),\, w \in \mathcal{W} \right\},
\end{equation*}
\vspace*{-.0cm}
and safety is guaranteed if $\mathcal{R}_k(\boldsymbol{\theta}) \subseteq \mathcal{X}_{\text{safe}}$ for all $k$. The key observation is that uncertainty in different state components contributes unequally to the growth of $\mathcal{R}_k$. This sensitivity is quantified by evaluating how perturbations in individual estimation errors $e^{(i)}$ affect safety-relevant metrics, such as deviation from a nominal trajectory or distance to constraint boundaries. Formally, a sensitivity score $s_i$ is computed for each state component by measuring the marginal increase in reachable set volume or safety margin violation when only the corresponding uncertainty bound is increased. The GPU partitioning problem is then posed as a constrained optimization:
\begin{equation*}
    \min_{\boldsymbol{\theta}} \; C(\boldsymbol{\theta}) 
    \quad \text{s.t.} \quad 
    \mathcal{R}_k(\boldsymbol{\theta}) \subseteq \mathcal{X}_{\text{safe}} \;\; \forall k,
\end{equation*}
where $C(\boldsymbol{\theta})$ models GPU cost (e.g., execution time or memory footprint). Rather than solving this problem exhaustively, the proposed heuristics exploit the sensitivity structure to allocate larger DNNs to perception tasks with high $s_i$, while aggressively shrinking models for low-impact state variables. This yields substantial reductions in GPU usage while preserving system-level safety guarantees, demonstrating that safety-aware allocation of imperfect perception models is more effective than uniform accuracy maximization.

This methodology was implemented and evaluated on the F1/10 autonomous racing platform using the F1TENTH Gym simulator. The simulator models vehicle dynamics using a bicycle model and provides realistic LiDAR-based perception for closed-loop control under varying track geometries. Multiple DNN-based perception modules were executed concurrently on a shared GPU budget, each estimating different state components such as lateral displacement, heading error, and track width. Safety was evaluated by analyzing trajectory deviation and constraint violations under different GPU allocations, allowing a direct comparison between uniform and safety-driven DNN sizing strategies. The experiments demonstrated that allocating larger DNNs to safety-critical perception tasks---such as lateral displacement and track width estimation---resulted in significantly lower trajectory deviation and improved robustness, even when the total GPU budget was smaller than that of poorly allocated configurations. These results highlight that correct \emph{allocation} of imperfect ML components can be more important than simply increasing overall model size.

\subsection{Summary}

Together, these two examples demonstrated that the presence of imperfect ML components does not preclude the design of safe, reliable, and certifiable autonomous systems. By explicitly modeling inference uncertainty, computation delays, and resource constraints at the system level, it is possible to analyze and reason~\cite{MaKCT04} about their combined impact on closed-loop behavior and to design control and perception architectures~\cite{GoSC11b}, and scheduling~\cite{XuGHTC23} and synthesis strategies~\cite{YeSVC25} that satisfy quantitative safety properties. This system-level perspective enables principled design decisions for edge--cloud partitioning and GPU resource allocation, moving beyond ad-hoc deployment of ML components. As a result, autonomous CPS with embedded DNN-based perception can achieve dependable operation and support certification, even in the presence of unavoidable imperfections \cite{capogrosso2024mtl}. 

While we have discussed how imperfect ML components, when suitably integrated inside a CPS, can still guarantee safety, similar ideas can be extended to security. For example, instead of authenticating~\cite{MuSLFC15b} or monitoring~\cite{WaMSLKC17} all the messages in the system, they could only be partially authenticated. If there is no attack on the system then this obviously leads to no change in system-level performance. But in the event of an attack---that can only impact the unauthenticated messages or portions of them---the system-level behavior can only change to a limited extent. This may be viewed as a loss in performance, but not a safety violation. Given a system-level safety specification, the problem here is to identify exactly \emph{how much} protection (e.g., authentication or monitoring) is needed to ensure that potential attacks on the system can deviate the closed-loop system trajectory (in its state space) not by too much and its specified safety property remains satisfied. In this process, by saving the amount of protection that is needed, a more resource-efficient design would become possible.


\section{Challenges and Regulations in Reliable and Secure AI Sensing Systems}
\label{sec:SecureAI}
AI-driven sensing systems are rapidly moving from labs to the edge—into environmental monitoring stations, radar units, vehicles, and human–machine interfaces. In safety- and mission-critical contexts, user trust is conditioned on demonstrable reliability, security, and clarity of purpose. This section proposes a practical roadmap for engineering trustworthy, user‑centric AI at the edge that aligns with evolving regulation while meeting real-world constraints. 

\subsection{Regulatory landscape and implications for edge sensing}

The EU Artificial Intelligence Act is the first comprehensive legal framework for AI, entering into force in August 2024 and establishing uniform, risk based obligations across the EU~\cite{Ref1}. It mandates an advisory forum to involve stakeholders and provide technical expertise to the Board and Commission, and tasks CEN and CENELEC with translating the Act’s principles into concrete technical requirements~\cite{Ref2}. In support of implementation, CEN CLC/JTC 21 is developing European standards intended to grant manufacturers a presumption of conformity, with actionable tests and benchmarks still being developed. 
In parallel, organizational best practices for trustworthy AI—transparency, human oversight, and codes of conduct—remain essential even for limited- and low-risk systems, and their proactive adoption can create competitive advantages and unlock innovation.

For automotive, the regulatory and assurance expectations intersect with cybersecurity and functional safety considerations. Connected vehicles must address V2X and vehicle control security to protect safety-critical functions and maintain resilience under adversarial conditions. More broadly, engineering methods must bridge the gap between imperfect learning components and certifiable system-level dependability to enable safety cases and homologation.

\begin{figure*}
    \centering
    \includegraphics[width=0.92\linewidth]{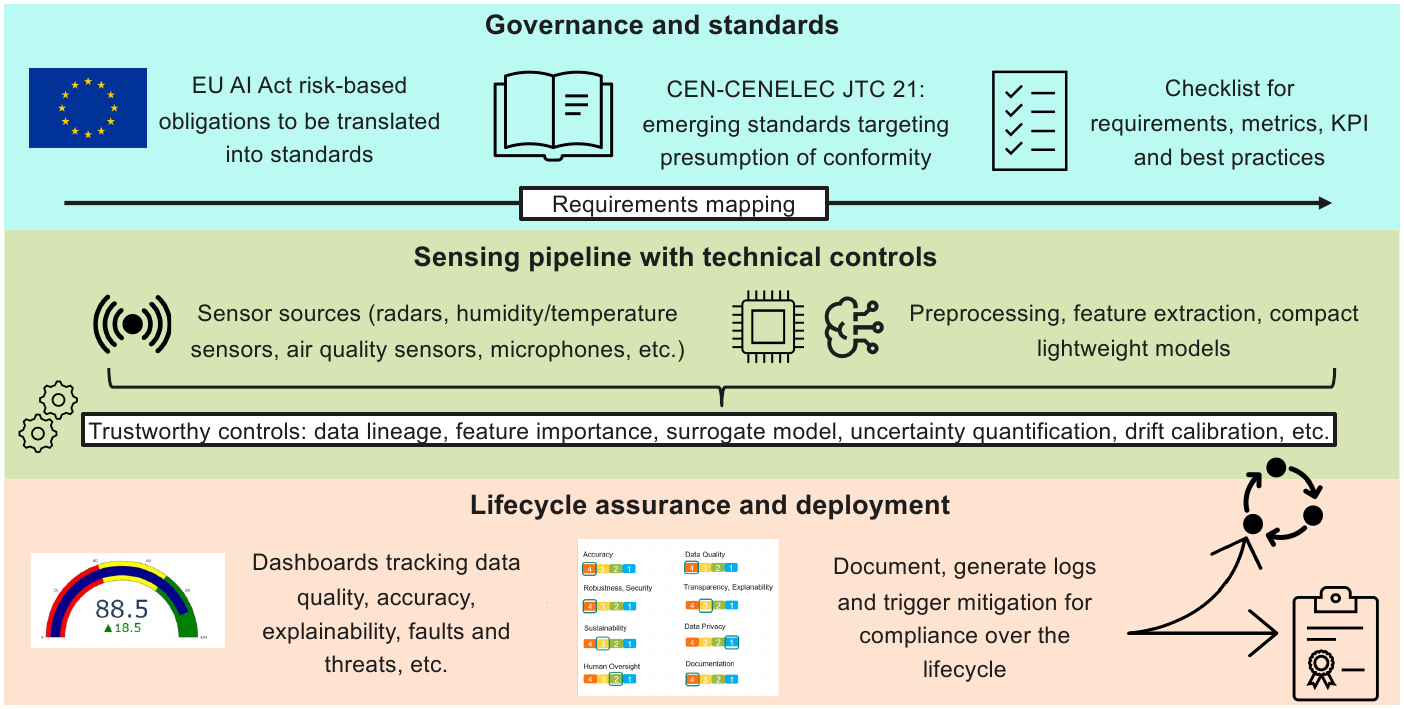}
    \vspace*{-.1cm}
    \caption{Trustworthy Edge AI Sensing: from Regulations to Technical Benchmarking
}
    \label{fig:TrustworthyAI}
    \vspace*{-.25cm}
\end{figure*}

\subsection{Engineering roadmap for trustworthy edge AI sensing}

The roadmap centers on six pillars that operationalize compliance, transparency, and reliability in edge deployments:

\subsubsection{Governance and requirement mapping}
Define scope, purpose, and intended use, including foreseeable misuse and faults and operational design domain. Identify risk levels and obligations and capture human oversight requirements early to prepare fallback strategies. For alignment with the EU AI Act, track emerging JTC 21 standards that provide presumption of conformity and translate obligations into tests. Even for limited- and low-risk functionality, implement transparency and oversight measures as baseline guardrails to gain strategic benefits.

\subsubsection{Data and pipeline integrity}
Institute dataset lineage, versioning, and audit trails across collection, curation, augmentation, and labeling. Incorporate environmental and device-level health checks (e.g., sensor drift, temperature, interference) and build redundancy where feasible. Use synthetic data, controlled perturbations and data augmentation to probe corner cases without exceeding edge resource budgets. Establish coverage metrics that reflect the operational design domain and known hazards.

\subsubsection{Reliability through lightweight uncertainty quantification and calibration}
Deploy uncertainty strategies such as Monte Carlo dropout and shallow ensembles when feasible and lightweight methods such as evidential deep learning~\cite{Mittermaier2023} for more resource constrained applications. Couple uncertainty estimates to downstream decision thresholds and user feedback. Maintain calibration under domain and concept shifts via periodic recalibration and small-footprint adaptation and identify out-of-distribution inputs early to trigger safe fallback behavior.

\subsubsection{Interpretability and rule extraction across sensor pipelines}
Extract interpretable decision rules from trained models via distillation and proxy surrogate modeling. Combine post-hoc model explanations with device-level feature attribution (e.g., feature importance ranking, inspection of attention layers, counterfactual examples, class activation mapping, etc.). Provide human-understandable rationales that increase operator and user trust. Lifecycle monitoring and documentation should explicitly include accuracy and intervenability measures, enabling corrective action when performance degrades~\cite{seifi2025complying}.

\subsubsection{Security and resilience-by-design}
Conduct threat and vulnerability analysis that spans the entire sensing pipeline, from raw signals to fused decisions, and includes cyber-physical attack vectors (spoofing, jamming, signal injection), data poisoning, and adversarial perturbations. In connected vehicles, explicitly address V2X security and vehicle control layer security to protect critical interfaces and ensure graceful degradation. Combine robust training with runtime anomaly detection and integrity checks to safeguard the model and its inputs. Maintain secure boots, signed models, and tamper-evident logs, and integrate with incident response processes.

\subsubsection{Lifecycle operations, monitoring, and continuous compliance}
Establish performance baselines and target ranges for data quality, accuracy, calibration, and uncertainty under representative scenarios that reflect environmental shifts and hardware aging. Monitor and shape accuracy metrics over the lifecycle, and track technical measures that support intervenability and oversight, aligning with emerging standardization guidance. Automate documentation and report necessary KPIs for compliance and traceability, including risk assessment updates and model change logs.

\subsection{Building organizational competence and AI literacy}
As requirements and standards remain in flux, cultivating AI literacy across roles—engineering, product, legal, quality, and operations—is essential. We recommend:
\begin{itemize}
    \item Role-specific training on risk categories and related obligations, and oversight expectations in the relevant regulation context.
    \item Guidebooks that translate governance requirements into concrete engineering patterns and checklists (documentation, traceability, monitoring).
    \item Cross-functional model review boards that assess accuracy, robustness, transparency, and alongside security and safety analyses.
    \item Lifecycle metrics dashboards to track metrics and KPIs and support compliance reporting and intervention via mitigation measures.
\end{itemize}

\subsection{Concluding Remarks}
In conclusion, building trustworthy, user-centric AI at the edge for safety- and mission-critical sensing applications requires a practical and actionable roadmap that prioritizes compliance, transparency, and reliability. By adopting trustworthy AI engineering best practices and anchoring them to relevant regulations, organizations can accelerate innovation, enable new business opportunities, and strengthen market differentiation. 

The automotive industry is a prime example of the need for trustworthy AI engineering. Edge AI powers ADAS and in-cabin sensing under stringent functional-safety and cybersecurity expectations. Uncertainty-aware, interpretable, and well-calibrated sensor pipelines are essential for safety cases, homologation, and long-term reliability in harsh operating conditions. By prioritizing trustworthy AI engineering, automotive manufacturers can ensure compliance with evolving regulations, such as the EU AI Act, and deliver strategic benefits that accelerate innovation and unlock new business opportunities.


\section*{Acknowledgments}

This research was supported in part by Deutsche Forschungsgemeinschaft~(DFG) through the Project \textit{LeanMICS} under Grant 534919862. Chakraborty's research was funded by the NSF grant 2038960 and a Dieter Schwarz Courageous Research Grant from Germany. 



\balance

\bibliographystyle{IEEEtran} 
\bibliography{Ref.bib,chakraborty}

\end{document}